\documentclass[letterpaper, 10 pt, conference]{ieeeconf}
\IEEEoverridecommandlockouts
\usepackage{cite}
\usepackage{amsmath,amssymb,amsfonts}
\usepackage{graphicx}
\usepackage{textcomp}
\usepackage{xcolor}
\usepackage{comment}
\usepackage{bm}
\usepackage{array}

\title{\LARGE \bf
Experimental Comparison of Whole-Body Control Formulations for Humanoid Robots in Task Acceleration and Task Force Spaces
}

\author{Sait Sovukluk$^{1, \dagger,*}$, Grazia Zambella$^{1,*}$, Tobias Egle$^{1}$, and Christian Ott$^{1,2}$
\thanks{$^{\dagger}$Corresponding author
({\tt\small sovukluk@acin.tuwien.ac.at})}
\thanks{$^{*}$Equal contribution in experiments and implementation.}
\thanks{$^{1}$Automation and Control Inst. (ACIN), TU Wien, 1040 Vienna, Austria}%
\thanks{$^{2}$Institute of Robotics and Mechatronics, German Aerospace Center (DLR), 82234 Weßling, Germany}%
\thanks{This research has received funding from the European Research Council (ERC) under the European Union’s Horizon 2020 research and innovation programme (grant agreement No. 819358) and the Austrian Academy of Sciences ({\"O}AW) and the Austrian Science Fund (FWF) under the Disruptive Innovation - Early Career Seed Money funding program (project HiFliTE).}
}

\begin{document}
\maketitle
\thispagestyle{empty}
\pagestyle{empty}

\begin{abstract}
This paper studies the experimental comparison of two different whole-body control formulations for humanoid robots: inverse dynamics whole-body control (ID-WBC) and passivity-based whole-body control (PB-WBC). The two controllers fundamentally differ from each other as the first is formulated in task acceleration space and the latter is in task force space with passivity considerations. Even though both control methods predict stability under ideal conditions in closed-loop dynamics, their robustness against joint friction, sensor noise, unmodeled external disturbances, and non-perfect contact conditions is not evident. Therefore, we analyze and experimentally compare the two controllers on a humanoid robot platform through swing foot position and orientation control, squatting with and without unmodeled additional weights, and jumping. We also relate the observed performance and characteristic differences with the controller formulations and highlight each controller's advantages and disadvantages.
\end{abstract}
\section{Introduction}
Humanoid robots are multi-purpose complex systems that are designed to imitate human appearance and behavior \cite{humanoids, humanoids2}. They are designed to perform human labor tasks, such as carrying weights, manipulating the environment, performing high-risk tasks, and assisting/supporting humans in their environments. The DARPA Robotics Challenge in 2015 summarizes the best usage of such robots \cite{darpa1,darpa2}. The challenge required humanoid robots to replace humans in a dangerous rescue mission. This mission included driving a golf cart; walking through doors, uneven/unorganized surfaces, and stairs; using power tools such as drillers and cutters; and manipulating the environment such as opening/closing valves, replacing items, plugging in power cords, and opening doors.

Averaging more than 25 degrees of freedom (DoF) per robot \cite{humanoids}, humanoid robots include multiple sub-systems (arms, legs, torso, and head), all of which are dynamically and kinematically coupled. Humanoid robots are categorized as floating (non-fixed) base systems and require active balancing on their feet both to stand and navigate around. The remaining degree of freedom, e.g., arms, is then used to manipulate the environment to perform useful tasks.
\begin{figure}[t!]
\centerline{\includegraphics[width=0.85\columnwidth]{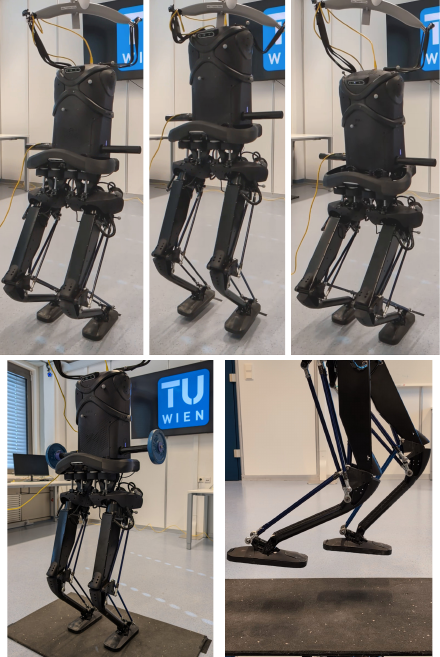}}
\caption{Snapshots of experiments covered in this paper: jumping (top), disturbed and undisturbed squatting (bottom left), and foot position and orientation control (bottom right). Supplemental video collects all experiments.}
\label{firstPageImg}
\end{figure}

The multi-task capability requires a controller for each task, called a task controller. The tasks may be keeping the center of mass (CoM) inside the support polygon, keeping the torso upright, applying a certain force at a certain point in task space through one of the arms, and carrying an item with the other arm. Then, all the tasks are combined in a framework called whole-body control (WBC), which solves the required motor inputs that satisfy all the tasks the best. The tasks can be combined and solved altogether with soft prioritization, allowing one task to affect the other tasks if they are not dynamically decoupled. In such formulation, task prioritization is achieved through weight selections such that a task with a higher weight affects the cost function more and results in higher prioritization. Strict prioritization, on the other hand, involves projecting the low-priority tasks into the nullspace of the higher-priority tasks such that the low-priority tasks cannot disturb the high-priority tasks. In such cases, the low-priority tasks are satisfied only if there are enough degrees of freedom left in the nullspace.

The literature collects multiple whole-body control frameworks with different task spaces and prioritization methods. Operational space whole-body control \cite{operationSpaceControl, sentis2005synthesis} formulates a whole-body control in task force space through strict prioritization. The inverse dynamics whole-body control (ID-WBC) is a combination of multiple task acceleration space controllers with soft \cite{IDC1-soft, IDC2-soft, IDC-imp} and strict \cite{IDC3-strict, kino} prioritization. The passivity-based whole-body control (PB-WBC) is a combination of passivity-based task space controllers in task force space with strict or soft prioritization \cite{mptc, PB-WBC, passivity1}.

This study compares two fundamentally different whole-body controllers: inverse dynamics whole-body control (ID-WBC) and passivity-based whole-body control (PB-WBC). ID-WBC is in task acceleration space and hence includes inertia shaping for each task definition. PB-WBC, on the other hand, is defined in task force space and does not include inertia shaping. Both controllers inherit stable closed-loop characteristics under ideal conditions, that is, without any disturbance, modeling errors, sensor noise, and joint friction. On the other hand, the methods' ability to handle real-life conditions remains uncertain. In this study, we design four different experimental scenarios:
\begin{enumerate}
    \item Swing foot position and orientation control (while the robot is hanging from a crane) experiment tests each controller's robustness against friction and modeling errors for low inertia tasks.
    \item Squatting experiment tests each controller's robustness against friction and modeling errors for high inertia tasks, i.e., CoM and body orientation control tasks.
    \item Squatting with an unmodeled weight experiment compares the robustness against disturbances through the same parameters tuned for the undisturbed case.
    \item Lastly, jumping experiment. As jumping goes through stance-flight-stance phases, it allows us to determine the robustness of each controller against hybrid dynamics, contact transitions, and imperfect contact conditions, as there will be a delay between the actual and sensed contact resulting in impact.
\end{enumerate}
The content of the paper follows:
\begin{itemize}
    \item Backgrounds on the inverse dynamics whole-body control (ID-WBC) and passivity-based whole-body control (PB-WBC) formulations appearing in the literature.
    \item An analysis on characteristic differences between two formulations and how external disturbances appear in different control spaces.
    \item Experimental results and discussion section collecting the experimental data, comparison, and discussion on the match between the analyses and experiments.
    \item Finally, a conclusion that highlights each controller's advantages and disadvantages for different scenarios.
\end{itemize}
\section{Background on ID-WBC Formulation}
This section provides a background on the inverse dynamics whole-body control formulation \cite{IDC-imp,IDC1-soft,IDC2-soft,IDC3-strict}. The details of the formulation will be relevant and useful when discussing the analytical and experimental comparison and implementation details.
\subsection{System Dynamics}
Let $\bm{q}$ be a set of configuration variables and $\bm{\nu} = (\bm{\nu}_{b}, \bm{\nu}_{j})$ be the generalized velocity where $\bm{\nu}_{b} = (\bm{v}_{b},\bm{\omega}_{b}) \in \mathbb{R}^{6}$ is the linear and angular velocity of the floating base and $\bm{\nu}_{j} \in \mathbb{R}^{n}$ is the generalized velocity of the joints. The well-known floating base robotic system dynamics results in
\begin{multline} \label{systemDyn}
\bm{M}(\bm{q}) \dot{\bm{\nu}}
+ \underbrace{\bm{C}(\bm{q},\bm{\nu})\bm{\nu} + \bm{\tau}_{g}(\bm{q})}_{\bm{h}(\bm{q},\bm{\nu})} = 
\underbrace{
\begin{bmatrix}
    \bm{0}\\
    \bm{\tau}
\end{bmatrix}
+
\bm{J}_{c}(\bm{q})^{\top} \bm{f}_{c}}_{\bm{\tau}_\text{cont}}\\
+
\underbrace{
\begin{bmatrix}
    \bm{0}\\
    \bm{\tau}_{ext}
\end{bmatrix}
+
\bm{J}_{ext}(\bm{q})^{\top} \bm{f}_{ext}}_{\bm{\tau}_\text{dist}}
\end{multline}
where $\{\bm{M}, \bm{C}\} \in \mathbb{R}^{(n+6) \times (n+6)}$ are the inertia and Coriolis matrices; $\bm{\tau}_{g} \in \mathbb{R}^{n+6}$ is the gravitational vector; $\bm{\tau} \in \mathbb{R}^{n}$ is the joint torques and forces; $\bm{f}_{c}\in\mathbb{R}^{6n_{c}}$ is the vector of contact wrenches for $n_{c}$ number of contacts; $\bm{\tau}_{ext} \in \mathbb{R}^{n}$ is the combination of all joint level disturbances such as friction and actuation uncertainties; $\bm{f}_{ext}$ is the combination of any number of task level disturbances and uncertainties; $\bm{J}_{(\cdot)}$ represent Jacobian matrix. Lastly, $\bm{\tau}_\text{dist}$ and $\bm{\tau}_\text{cont}$ are a combination of disturbance and control terms, respectively.
\subsection{Task Space Inverse Dynamics Control}
Let $\bm{J}_{i}$ be a Jacobian mapping between the joint space and the $i^{th}$ task space such that the task space velocity $\dot{\bm{x}}_{i}$ corresponds to
\begin{equation}
    \dot{\bm{x}}_{i} = \bm{J}_{i} \bm{\nu}.
\end{equation}
Similarly, the task space acceleration is given as
\begin{equation} \label{idc-taskAcc}
    \ddot{\bm{x}}_{i} = \bm{J}_{i}\dot{\bm{\nu}} +  \dot{\bm{J}}_{i}\bm{\nu}.
\end{equation}
Let
\begin{equation} \label{idc-desAcc}
    \ddot{\bm{x}}_{i,d} = \ddot{\bm{x}}_{i,\text{ref}} + \bm{K}_{d,i} \underbrace{(\dot{\bm{x}}_{i,\text{ref}} - \dot{\bm{x}}_{i})}_{\dot{\bm{e}}} + \bm{K}_{p,i} \underbrace{(\bm{x}_{i,\text{ref}} - \bm{x}_{i})}_{\bm{e}}
\end{equation}
be a desired task space acceleration where $\bm{K}_{p}$ and $\bm{K}_{d}$ are positive definite gain matrices. If there exists a feasible set of control parameters $\bm{\tau}_\text{cont}$ such that $\ddot{\bm{x}}_{i,d} = \ddot{\bm{x}}_{i}$ and $\bm{\tau}_\text{dist} = \bm{0}$, the resultant closed-loop dynamics is exponentially stable
\begin{equation} \label{idc-cld}
    \ddot{\bm{e}}_{i} + \bm{K}_{d,i} \dot{\bm{e}}_{i} + \bm{K}_{p,i} \bm{e} = \bm{0}.
\end{equation}
A combination of all task controllers, such as balancing and manipulation, constitute the whole-body controller. If the task mappings $\bm{J} = [\bm{J}_{1}; \bm{J}_{2}; \dots]$ is square and invertible, then the control system is well defined and the solution is unique. If underconstrained, the control system is partially stabilized. Finally, if overconstrained, the whole-body controller compromises between different tasks depending on the hierarchy or weight selection.
\subsection{Whole-Body Control Formulation}
The whole body control formulation gathers all task controls and combines them in a QP (quadratic problem):
\begin{subequations} \label{QPeqID}
    \begin{equation} \tag{\ref{QPeqID}}
    \underset{\dot{\bm{\nu}}, \bm{\tau}, \bm{f}_{c}}{min} \sum_{i}{(\ddot{\bm{x}}_{i,d} -\ddot{\bm{x}}_{i})^{\top} \bm{W}_{i}(\ddot{\bm{x}}_{i,d} - \ddot{\bm{x}}_{i})}
    \end{equation}
    \begin{equation*}
        \text{Such that:}
    \end{equation*}
    \begin{equation} \label{c1}
    \bm{M}\dot{\bm{\nu}} + \bm{C}\bm{\nu} + \bm{\tau}_{g} = \bm{\tau}_\text{cont}
    \end{equation}
    \begin{equation} \label{c2}
    \bm{J}_{c}\dot{\bm{\nu}} +  \dot{\bm{J}}_{c}\bm{\nu} = 0
    \end{equation}
    \begin{equation} \label{c3}
    \left| f_{x,l} \right| \leq \dfrac{\mu f_{z,l}}{\sqrt{2}}\text{,} \enspace \left| f_{y,l} \right| \leq \dfrac{\mu f_{z,l}}{\sqrt{2}}\text{,} \enspace \text{and} \enspace f_{z} \geq 0 \enspace \forall l
    \end{equation}
    \begin{equation} \label{c4}
    \bm{\tau}_\text{min} \leq \bm{\tau} \leq \bm{\tau}_\text{max}
    \end{equation}
\end{subequations}
where $\bm{W}_{i}$ are the corresponding weight matrices in a soft prioritization form; \eqref{c2} is contact constraint; \eqref{c3} is friction constraint for active contacts; $\bm{\tau}_\text{min}$ and $\bm{\tau}_\text{max}$ are the minimum and maximum limits of input torques and forces. A detailed derivation of ID-WBC is given in \cite{IDC2-soft, IDC3-strict} and for systems with closed kinematic chains in \cite{IDC1-soft}.
\section{Background on PB-WBC Formulation}
This section provides a background on the passivity-based whole-body control \cite{PB-WBC,PB-WBC-imp}. The details of the formulation will be relevant and useful when discussing the analytical and experimental comparison and implementation details.
\subsection{System Dynamics}
In the case of PB-WBC, the translational base position and velocity states are replaced by the CoM position and velocity states through a CoM frame located at the CoM and aligned with the inertial frame \cite{passivityDynamics,PB-WBC,PB-WBC-imp}. Consequently, redefine the generalized velocity $\bm{\nu}$ as $\bm{\nu}_{p} = (\bm{\nu}_{c}, \bm{\nu}_{j})$ where $\bm{\nu}_{c} = (\bm{v}_{c},\bm{\omega}_{b}) \in \mathbb{R}^{6}$ is the linear velocity of CoM and angular velocity of the floating base and $\bm{\nu}_{j} \in \mathbb{R}^{n}$ is the generalized velocity of the joints. Similarly, redefine the inertia $\bm{M}$ and Coriolis $\bm{C}$ matrices as $\{\bm{M}_{c}, \bm{C}_{c}\}$ where the translational portion of the floating base is not w.r.t. the base frame but the CoM frame. Consequently, the gravitational vector $\bm{\tau}_{g}$ takes a special form
\begin{equation*}
    \bm{\tau}_{g} =
    \begin{bmatrix}
        -\bm{w}_{g}\\
        \bm{0}_{n\times 1}
    \end{bmatrix},
    \enspace \text{where \enspace}
    \bm{w}_{g} = 
    \begin{bmatrix}
        m_{c}\bm{g}\\
        \bm{0}_{3\times 1}\\
    \end{bmatrix},
\end{equation*}
$m_{c} \in \mathbb{R}$ is the total mass, and $\bm{g} \in \mathbb{R}^{3}$ is the gravitational acceleration vector.
\subsection{Task Space Passivity-Based Control}
Unlike ID-WBC, the PB-WBC assumes a well-defined system: the combination of task mappings $\bm{J}=[\bm{J}_{1};\bm{J}_{2};\dots]$ is square, invertible, and covers all DoF \cite{PB-WBC}. In the context of legged locomotion, the overall task space mapping is a combination of some logical tasks \cite{PB-WBC}: a 6-dimensional tracking task for the CoM states ($\bm{v}_{c} \in \mathbb{R}^{3}$) and torso orientation ($\bm{\omega}_{b} \in \mathbb{R}^{3}$); a contact constraint task that keeps the contact points stationary and generates the resultant interaction and reaction forces; a combination of impedance tracking tasks for the remaining DoF such as arms and leg that are not in contact with the ground or environment. Consequently,
\begin{equation} \label{generalVelPB}
    \underbrace{
    \begin{bmatrix}
        \bm{\nu}_{c} \\
        \dot{\bm{x}}_\text{grf}\\
        \dot{\bm{x}}_\text{imp}
    \end{bmatrix}}_{\dot{\bm{x}}}
    =
    \underbrace{
    \begin{bmatrix}
        \bm{I}_{6 \times 6} & \bm{0}_{6 \times n} \\
        \bm{J}_{\text{grf,c}} & \bm{J}_{\text{grf,j}}\\
        \bm{J}_{\text{imp,c}} & \bm{J}_{\text{imp,j}}
    \end{bmatrix}}_{\bm{J}}
    \begin{bmatrix}
        \bm{\nu}_{c} \\
        \bm{\nu}_{j}
    \end{bmatrix},
\end{equation}
where $\dot{\bm{x}} \in \mathbb{R}^{n+6}$ is combination of generalized task velocities; $\bm{J} \in \mathbb{R}^{(n+6) \times (n+6)}$ is the combination of Jacobian mappings; $(\cdot)_\text{grf}$ is the contact task for the stationary task frame;  $(\cdot)_\text{imp}$ is the impedance tasks for the remaining degree of freedom, respectively. The desired generalized velocities are related to the desired task space velocities through
\begin{equation} \label{pb_jointDes}
    \begin{bmatrix}
        \bm{\nu}_{c,d}\\
        \bm{\nu}_{j,d}
    \end{bmatrix}
    =
    \bm{J}^{-1} \dot{\bm{x}}_{d}.
\end{equation}
For an error definition of $\tilde{\bm{\nu}} = \bm{\nu}_{d} - \bm{\nu}$, the desired PD+ \cite{pd+} alike closed-loop dynamics is written as
\begin{equation} \label{P-CLD}
    \bm{M}_{c}
    \begin{bmatrix}
        \dot{\tilde{\bm{\nu}}}_{c}\\
        \dot{\tilde{\bm{\nu}}}_{j}
    \end{bmatrix}
    +
    \bm{C}_{c}
    \begin{bmatrix}
        \tilde{\bm{\nu}}_{c}\\
        \tilde{\bm{\nu}}_{j}
    \end{bmatrix}
    =
    -\bm{J}^{\top}
    \begin{bmatrix}
        \bm{w}_{c}^\text{imp}\\
        \bm{f}_\text{grf}\\
        \bm{f}_\text{imp}
    \end{bmatrix}
    - \bm{\tau}_\text{dist},
\end{equation}
where $\bm{w}_{c}^\text{imp} \in \mathbb{R}^{6}$ is the desired impedance wrench for the CoM and body orientation states, $\bm{f}_\text{grf}$ is the resultant contact forces at the contact frames, and $\bm{f}_\text{imp}$ is the desired impedance task force for the remaining DoF \cite{PB-WBC}.
\subsection{Whole-Body Control Formulation}
The substitution of the desired closed-loop dynamics \eqref{P-CLD} into the system dynamics results in
\begin{equation} \label{P-eqn} \scalebox{0.94}{$
    \begin{bmatrix}
        \bm{M}_{c,1}\\
        \bm{M}_{c,2}\\
    \end{bmatrix}
    \begin{bmatrix}
        \dot{\bm{\nu}}_{c,d}\\
        \dot{\bm{\nu}}_{j,d}\\
    \end{bmatrix}
    +
    \begin{bmatrix}
        \bm{C}_{c,1}\\
        \bm{C}_{c,2}\\
    \end{bmatrix}
    \begin{bmatrix}
        \bm{\nu}_{c,d}\\
        \bm{\nu}_{j,d}\\
    \end{bmatrix}
    =
    \begin{bmatrix}
        \bm{w}_{g}\\
        \bm{\tau}
    \end{bmatrix}
    +
    \bm{J}^{\top}
    \begin{bmatrix}
        \bm{w}_{c}^\text{imp}\\
        \bm{f}_\text{grf}\\
        \bm{f}_\text{imp}
    \end{bmatrix}
    $},
\end{equation}
where matrix subscripts $(\cdot)_{1}$ and $(\cdot)_{2}$ indicates the first six and the last $n$ rows, respectively. The feasible set of ground reaction forces is decoupled from the input torques, hence, can be solved through the first six rows of \eqref{P-eqn}:
\begin{subequations} \label{QPeqPB}
\begin{equation}
    \min_{\bm{f}_\text{grf}}\left( \bm{\delta}_{c}^{\top}\bm{Q}_{c} \bm{\delta}_{c} + \bm{\delta}_{f}^{\top}\bm{Q}_{f} \bm{\delta}_{f} \right)
\end{equation}
\begin{equation*}
    \text{Such that:}
\end{equation*}
\begin{equation}\label{first6}\scalebox{0.89}{$
    \bm{\delta}_{c} = \bm{J}_\text{grf,c}^{\top} \bm{f}_\text{grf} - \bm{M}_{1}\dot{\bm{\nu}}_{d} - \bm{C}_{1}\bm{\nu}_{d} + \bm{w}_{g} + \bm{w}_{c}^\text{imp} + \bm{J}_\text{imp,c}^{\top} \bm{f}_\text{imp}
    $}
\end{equation}
\begin{equation}
    \bm{\delta}_{f} = \bm{f}_\text{grf,d} - \bm{f}_\text{grf}
\end{equation}
\begin{equation}
    \text{feasibility constraints: friction, bounds, \dots}
\end{equation}
\end{subequations}
where \eqref{first6} is the first six rows of \eqref{P-eqn}; $\bm{f}_\text{grf,d}$ is the combination of desired reaction forces, if any, originating from the contact points; $\bm{Q}_{c}$ and $\bm{Q}_{f}$ are the weight matrices in a soft prioritization form between the CoM task and the desired ground reaction force task if any. Note that zero $\bm{f}_\text{grf,d}$ with a positive definite $\bm{Q}_{f}$ is a damping task on the ground reaction forces. The resultant joint control torques and forces are then calculated through the last $n$ rows of \eqref{P-eqn},
\begin{equation} \label{pb_torque}
    \bm{\tau}_\text{PBC} = \bm{M}_{2}\dot{\bm{\nu}}_{d} + \bm{C}_{2}\bm{\nu}_{d} -
    \begin{bmatrix}
        \bm{J}_\text{grf,j}^{\top} & \bm{J}_\text{imp,j}^{\top}
    \end{bmatrix}
    \begin{bmatrix}
        \bm{f}_\text{grf}\\
        \bm{f}_\text{imp}
    \end{bmatrix}.
\end{equation}
\section{Closed-Loop Dynamics, Control Gains, and Disturbance Appearance Analysis}
As each controller is formulated in different task spaces, the PD (proportional-derivative) control actions and error scaling through these control gains result in different behaviors. Furthermore, the external joint and task level disturbances, $\bm{\tau}_\text{dist}$, appear and scale differently in different task spaces. This section analyses both control systems' behavioral and characteristic differences in terms of the control gains, error scaling, and external disturbance appearances.
\subsection{ID-WBC}
The required control action to achieve the desired closed-loop dynamics \eqref{idc-cld} can be calculated by substituting \eqref{systemDyn} into \eqref{idc-taskAcc} for any task control,
\begin{equation}
    \ddot{\bm{x}} = \bm{J}\bm{M}^{-1}(\bm{\tau}_\text{cont} + \bm{\tau}_\text{dist} - \bm{h}) +  \dot{\bm{J}}\bm{\nu}.
\end{equation}
Setting the control action as
\begin{equation} \label{idc-cont}
    \bm{J}\bm{M}^{-1}\bm{\tau}_\text{cont} = \bm{J}\bm{M}^{-1}\bm{h} - \dot{\bm{J}}\bm{\nu} + \ddot{\bm{x}}_{d},
\end{equation}
where $\ddot{\bm{x}}_{d}$ is given in \eqref{idc-desAcc} and $\bm{\tau}_\text{dist}$ is treated unknown during the control design, results in the given closed-loop dynamics
\begin{equation} \label{idc-actual}
    \ddot{\bm{e}} + \bm{K}_{d} \dot{\bm{e}} + \bm{K}_{p} \bm{e} = -\bm{J}\bm{M}^{-1} \bm{\tau}_\text{dist},
\end{equation}
where $\bm{\tau}_\text{dist}$ is the combination of any disturbances, such as friction, non-ideal actuator dynamics, and any unmodeled external force originating from the environment as shown in \eqref{systemDyn}. In case $\bm{\tau}_\text{dist} = \bm{0}$ the exponential stability is obvious.

Two important observations can be made through \eqref{idc-cont} and \eqref{idc-actual}. The first observation from \eqref{idc-cont} is the errors appearing in $\ddot{\bm{x}}_{d}$ are compensated through acceleration commands, that is, setting $\bm{K}_{p} = \bm{K}_{d} = \bm{I}$, a unit error in position and velocity results in a unit additional acceleration command. The resultant required torques and forces are scaled by the inertia matrix. The resultant force is scaled up for high inertia tasks, such as body position and orientation tasks, as a unit acceleration for high inertia task requires more force; scaled down for low inertia tasks, such as ankle position and orientation tasks. The same scaling also applies for disturbance rejection as the same amount of disturbance, for example, friction, requires a different acceleration effort depending on the inertia mapping. The second observation from \eqref{idc-actual} is the appearance of the unmodeled disturbances on the rhs (right-hand side) of the equation as \eqref{idc-actual} forms a second-order linear non-homogeneous ordinary differential equation. For the sake of analysis, let \eqref{idc-actual} be a 1-DoF model with a constant mass, disturbance, and control gains. The solution to the system,
\begin{multline} \label{ode}
    e(t) = c_{1} \exp{\left(\dfrac{-k_{d} + \sqrt{k_{d}^2 - 4k_{p}}}{2}t\right)} \\\ + c_{2} \exp{\left( \dfrac{-k_{d} - \sqrt{k_{d}^2 - 4k_{p}}}{2} t\right)} - \dfrac{\tau_\text{dist}}{mk_{p}},
\end{multline}
indicates a steady-state error determined by $\tau_\text{dist}/(mk_{p})$ term. Furthermore, as the given steady-state error is scaled by the inverse of mass, keeping $\bm{K}_{p}$ and $\bm{K}_{d}$ constant, the same amount of disturbance, for example, joint friction, results in different amounts of steady-state error for different tasks depending on the inertia. A high inertia task, such as CoM and body orientation control, results in a smaller steady-state error than a low inertia task, such as ankle position and orientation control. As a result, in the case of ID-WBC, the low-inertia tasks require higher control gains for the same amount of disturbance rejection as the high-inertia tasks.
\subsection{PB-WBC}
The same analysis for the passivity-based control appears to be more straightforward as the control actions and error definitions are in the same spaces. Setting the control action
\begin{equation} \label{pb-controlAction}
    \bm{\tau}_{\text{cont}} = 
    \bm{M}_{c} \dot{\bm{\nu}}_{d}
    + \bm{C}_{c} \bm{\nu}_{d} +
    \begin{bmatrix}
        -\bm{w}_{g}\\
        \bm{0}_{n \times 1}
    \end{bmatrix}
    + \bm{J}^{\top}
    \begin{bmatrix}
        \bm{w}_{c}^\text{imp}\\
        \bm{f}_\text{grf}\\
        \bm{f}_\text{imp}
    \end{bmatrix}
\end{equation}
and substituting it in the system dynamics results in the given closed-loop dynamics
\begin{equation} \label{pb-actual}
    \bm{M}_{c} \dot{\tilde{\bm{\nu}}}
    +
    \bm{C}_{c}  \tilde{\bm{\nu}} +
    \bm{J}^{\top}
    \begin{bmatrix}
        \bm{w}_{c}^\text{imp}\\
        \bm{f}_\text{grf}\\
        \bm{f}_\text{imp}
    \end{bmatrix}
    =
    - \bm{\tau}_\text{dist}.
\end{equation}
The first difference between the ID-WBC and PB-WBC appears in \eqref{pb-controlAction} as the errors appearing in the impedance control formulations, for example, $\bm{f}_\text{imp} = \bm{K}_{d}\tilde{\bm{v}}_{j} + \bm{K}_{p}\tilde{\bm{p}}_{j}$ for any task, is not scaled by the inertia matrix before converted into actions. This property eliminates the behavioral differences in different tasks depending on the inertia mapping. Similar observations are also valid for the closed-loop dynamics \eqref{pb-actual}. Again, for the sake of analysis, assume a 1-DoF system for \eqref{pb-actual}. The solution to the system,
\begin{multline} \label{ode2}
    e(t) = c_{1} \exp{\left(\dfrac{-k_{d} + \sqrt{k_{d}^2 - 4mk_{p}}}{2m}t\right)} \\\ + c_{2} \exp{\left( \dfrac{-k_{d} - \sqrt{k_{d}^2 - 4mk_{p}}}{2m} t\right)} - \dfrac{\tau_\text{dist}}{k_{p}},
\end{multline}
indicates a steady-state error determined by $\tau_\text{dist}/k_{p}$ term. Unlike ID-WBC, in the case of PB-WBC, the same control gains and amount of disturbance result in the same steady-state error for all tasks regardless of their inertia. Consequently, the PB-WBC requires higher gains for high-inertia and lower gains for low-inertia tasks to achieve the same disturbance rejection as the ID-WBC.
\section{Experimental Results and Discussion}
We design four experiment scenarios to compare the different aspects of the whole-body control formulations. These are foot control, squatting with and without unmodeled weights, and jumping. We use Kangaroo bipedal robot as testbed, ProxQP \cite{proxqp} to solve the QPs appearing in \eqref{QPeqID} and \eqref{QPeqPB}, and Pinocchio \cite{pinocchio} to compute the dynamics. We experimentally tune the control gain parameters for each task without any additional disturbances to the best of our knowledge. In case of additional unmodeled weight, we use the same parameters tuned for the undisturbed case.
\subsection{Foot Position and Orientation Control}
\begin{figure}[t!]
\centerline{\includegraphics[width=\columnwidth]{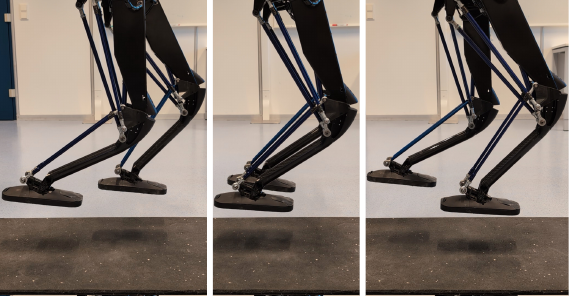}}
\caption{Snapshots of Kangaroo following the desired foot position and orientation trajectories while hanging from a crane.}
\label{hangingFrames}
\end{figure}

The foot control task is implemented through a foot frame per leg attached to the sole while the robot is hanging from a crane (see Fig.~\ref{hangingFrames}). The robot is commanded to swing the legs back and forward $15 cm$ in $x$ direction while keeping the height constant and leg orientation aligned with the inertial frame. Both controllers are tuned for the least steady-state error in position control without any oscillation.

\begin{figure}[t!]
\centerline{\includegraphics[width=\columnwidth]{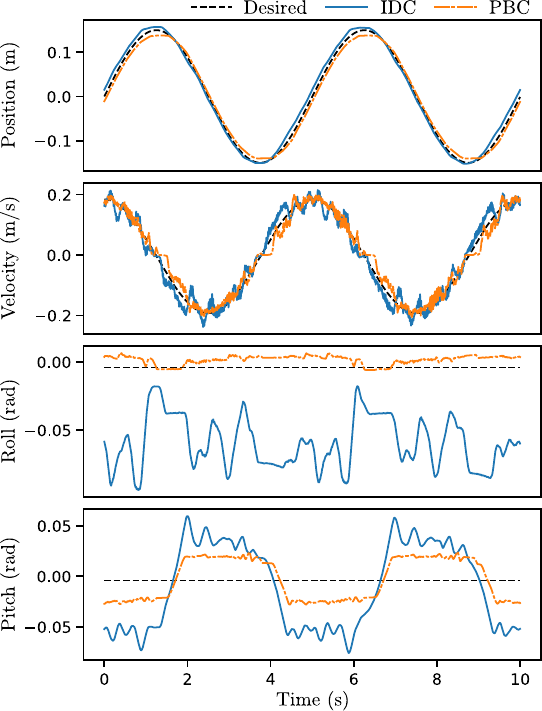}}
\caption{The response of ID-WBC and PB-WBC to the foot position and orientation control tasks. The feet are commanded to move back and forward at a constant height and orientation w.r.t. the inertial frame. The desired trajectory is a peak-to-peak $30cm$ sinusoidal signal with $0.2 Hz$ frequency.}
\label{hangingPlot}
\end{figure}

The Kangaroo robot suffers from modeling errors and friction in ankle joints originating from the closed-kinematic chains and distant motor placements to keep the legs lightweight. As the inverse dynamics control scales the disturbances with the inertia inverse, shown in \eqref{idc-actual} and \eqref{ode}, it requires high control gains for disturbed low-inertia systems. In the case of PB-WBC, however, the disturbances are not scaled, shown in \eqref{pb-actual} and \eqref{ode2}, and can be rejected with lower gains for low-inertia systems. The response of both controllers is shown in Fig.~\ref{hangingPlot}. The figure shows a similar control performance for both controllers with a slightly better orientation control in the case of PB-WBC. Due to the control space differences, ID-WBC gains are approximately a hundred times higher than the PB-WBC gains for the foot orientation control. A further increase in the gains results in marginal stability mostly due to the modeling errors and backlash. The PB-WBC is more intuitively tuned as its gains remain around a few hundred compared to the few tens of thousands in case ID-WBC due to the extremely low inertias appearing in the ankle joints.
\subsection{Squat Control}
\begin{figure}[t!]
\centerline{\includegraphics[width=\columnwidth]{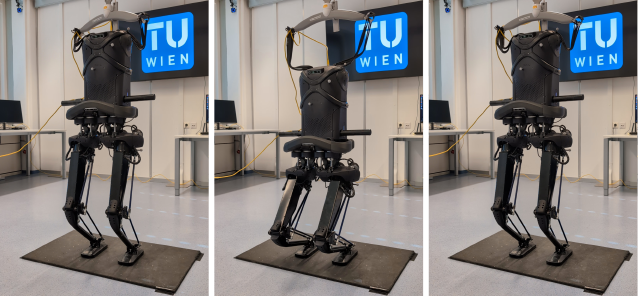}}
\caption{Snapshots of Kangaroo doing $20cm$ squats.}
\label{squatFrames}
\end{figure}

\begin{figure}[t!]
\centerline{\includegraphics[width=\columnwidth]{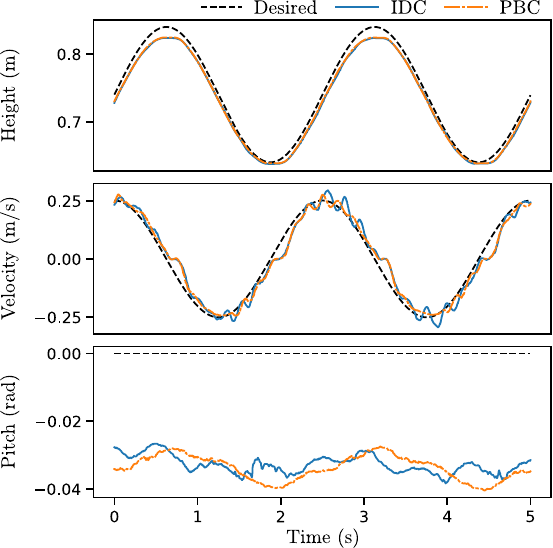}}
\caption{The response of ID-WBC and PB-WBC to squatting. The desired trajectory is a peak-to-peak $20cm$ sinusoidal signal with $0.4 Hz$ frequency.}
\label{squatPlot}
\end{figure}

Squatting involves multiple high-inertia tasks, which are CoM and torso orientation control. The snapshots of the robot performing squat are shown in Fig.~\ref{squatFrames}, where the desired trajectory is a peak-to-peak $20cm$ sinusoidal signal with $0.4Hz$ frequency. Both controllers are tuned for the least steady-state error in position without any oscillation. Additionally, with the advantage of the knowledge from \eqref{ode} and \eqref{ode2}, we scale the CoM height control gains of the ID-WBC by the total mass ($41$ $kg$) and use them in PB-WBC such that both controllers match in height control. The response of both controllers to the desired behavior is shown in Fig.~\ref{squatPlot}. Both control systems show remaining errors at the peak height. The anticipated disturbance in the CoM height control, that is,
\begin{equation*}
    \delta_\text{dist,ID} \approx \underbrace{(0.84 - 0.824)}_{\text{error}}\underbrace{(41)}_{m}\underbrace{(150)}_{k_{p}} = 98.4N
\end{equation*}
for ID-WBC through \eqref{ode} and 
\begin{equation*}
    \delta_\text{dist,PB} \approx \underbrace{(0.84 - 0.824)}_{\text{error}}\underbrace{(6100)}_{k_{p}} = 97.6N
\end{equation*}
for PB-WBC through \eqref{ode2}, indicates approximate $100N$ of disturbance combination for modeling errors, friction, and other uncertainties appearing in this task. Both methods perform a similar velocity tracking and torso orientation control performance with negligible differences.
\subsection{Squat Control with Additional Unmodeled Weight}
\begin{figure}[t!]
\centerline{\includegraphics[width=\columnwidth]{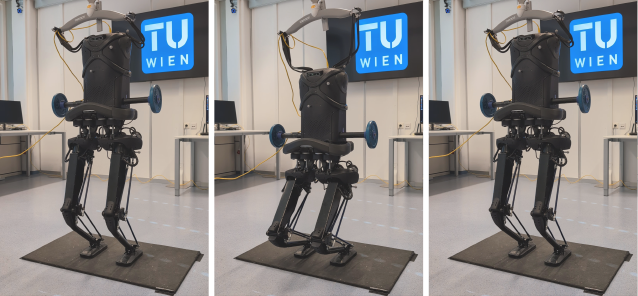}}
\caption{Snapshots of Kangaroo doing $20cm$ squats with $5kg$ of weights.}
\label{squatDistFrames}
\end{figure}

\begin{figure}[t!]
\centerline{\includegraphics[width=\columnwidth]{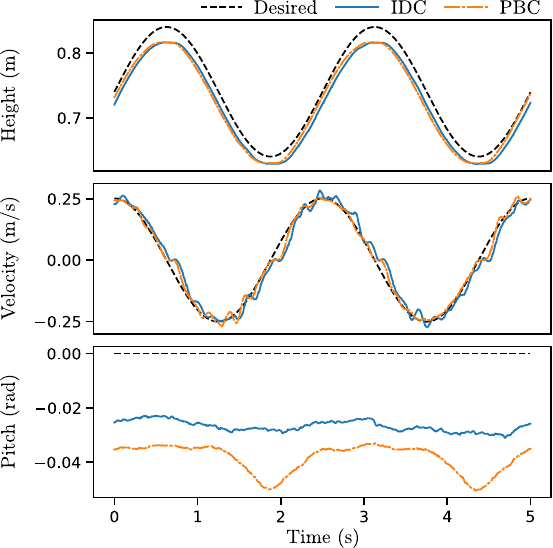}}
\caption{The response of ID-WBC and PB-WBC to squatting with an additional 5kg of unmodeled mass. The desired trajectory is a peak-to-peak 20cm sinusoidal signal with 0.4Hz frequency.}
\label{squatDistPlot}
\end{figure}

Squatting with additional unmodeled $5kg$ of mass (see Fig.~\ref{squatDistFrames}) is performed with the same control gains and desired trajectory as undisturbed squatting. The anticipated disturbance for $2.4$ $cm$ remaining error at the peak height $\delta_\text{dist,ID} \approx \delta_\text{dist,PB} \approx 147N$ matches with the previous observation as it is increased by the same amount as the additional weight. The response of both controllers (see Fig.~\ref{squatDistPlot}) shows an increase in steady-state error as expected.
\subsection{Jumping Control}
Jumping motion is generated through two trajectories for jumping and landing. First, a sixth-order polynomial is generated for jumping. The polynomial connects the initial height, velocity, and acceleration states to the desired jumping height, velocity, and acceleration states in $0.5$ seconds. At the moment of touchdown, again, a sixth-order polynomial is generated for landing in $0.5$ seconds. Online trajectory generation considering the system states both for jumping and landing results in smooth and continuous ground reaction forces. The liftoff is triggered once the CoM is above a threshold height with a positive velocity and the leg forces are zero. Similarly, the landing phase is triggered in the case of negative CoM velocity, zero feet height, and positive leg forces. The snapshots of jumping are provided in Fig.~\ref{firstPageImg}.

As the PB-WBC requires an invertible Jacobian and full state control as in \eqref{generalVelPB} and the body states cannot be controlled during the flight phase, we set desired body states to the current measurement, $\bm{\nu}_{c,d} = \bm{\nu}_{c}$, in \eqref{pb_jointDes}. Additionally, as mentioned in \cite{PB-WBC} for the PB-WBC, to prevent torque discontinuities at the touchdown moments, an additional damping factor is included for the stance phase such that $\bm{\tau}_\text{PBC}$ in \eqref{pb_torque} is rewritten as
\begin{equation} \label{damping}
    \bm{\tau}_\text{PBC, damped} =
    \begin{cases}
    \bm{\tau}_\text{PBC} - \bm{J}_\text{grf}^{\top}\bm{D}_\text{PBC}\dot{\bm{x}}_\text{grf}, &\enspace\enspace \text{landing}\\
    \bm{\tau}_\text{PBC}, \enspace &\text{jump \& flight}
    \end{cases}
\end{equation}
where $\bm{D}_\text{PBC} \in \mathbb{R}^{(6n_{c}) \times (6n_{c})}$ is a positive definite diagonal matrix for each contact point. This damping applies only if the rigid contact assumption is violated and the foot is not stationary on the ground. Furthermore, $\bm{Q}_{f}$ in the QP formulation \eqref{QPeqPB} is selected to be $10^{-5}$ for zero desired ground reaction forces $\bm{f}_\text{grf,d}$ in lateral directions in order to damp the lateral ground reaction forces.

The PB-WBC gains are obtained through scaling the ID-WBC gains, $41$ for the CoM task and $\approx5.5$ for the body orientation task, to achieve similar closed-loop dynamics and disturbance rejection performance as in the height control of squat experiments. The leg control task gains for the flight phase remain the same as the hanging experiment.
\begin{figure}[t!]
\centerline{\includegraphics[width=\columnwidth]{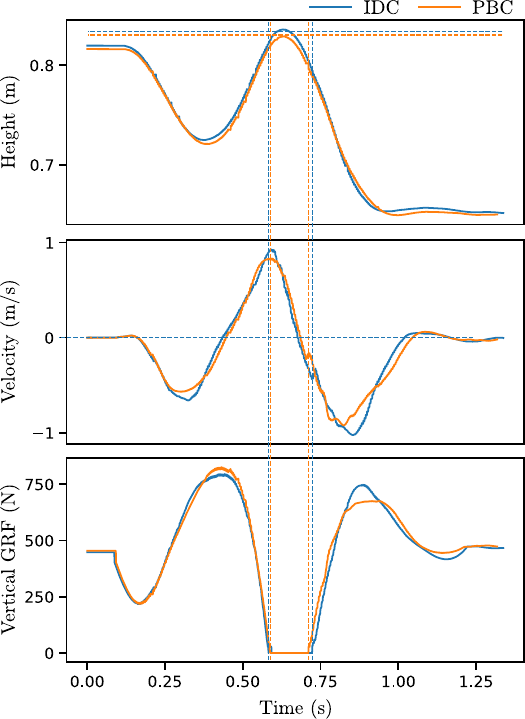}}
\caption{The response of ID-WBC and PB-WBC to $1.3$ $cm$ jump. The GRF represents the ground reaction force estimated by the whole-body controller. The vertical dashed lines represent the liftoff and touchdown moments. Horizontal lines represent the desired jumping height.}
\label{jump1Plot}
\end{figure}
\begin{figure}[t!]
\centerline{\includegraphics[width=\columnwidth]{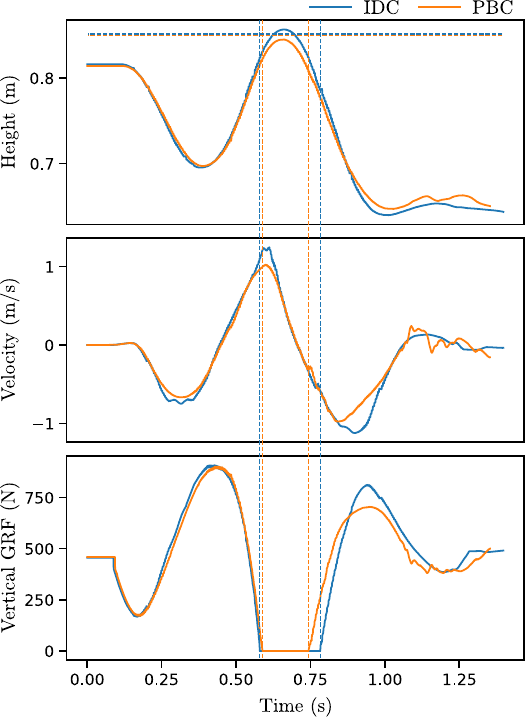}}
\caption{The response of ID-WBC and PB-WBC to $3$ $cm$ jump.}
\label{jump2Plot}
\end{figure}

We first perform a small jump shown in Fig.~\ref{jump1Plot}. The same standing height with both controllers indicates matching proportional control gains in height control. For this experiment, we set the additional damping $\bm{D}_\text{PBC} = \text{diag}(20,20,30,0,0,0)$ for each contact. When the same desired jumping behavior is applied, we observe $1$ $mm$ undershoot for the PBC and $2$ $mm$ overshoot for the IDC. The second jump with higher velocities is shown in Fig.~\ref{jump2Plot}. As the landing velocity increases, we update the additional damping values of PB-WBC to $\bm{D}_\text{PBC} = \text{diag}(20,30,40,0,0,0)$. When the same desired jumping behavior is applied, we observe $4$ $mm$ undershoot for the PBC and $6$ $mm$ overshoot for the IDC. A phase shift is apparent in both velocity data originating from the low-pass filter applied to the velocity estimation of the motion capture system.

When both jumping experiments in Fig.~\ref{jump1Plot} and Fig.~\ref{jump2Plot} are examined, small differences in velocity and ground reaction response are observed. These differences mostly originate from additional damping applied to the PBC, mismatches in overall system tuning, and differences in the closed-loop dynamics, for example, the Coriolis matrix remains in the closed-loop dynamics of PBC but not in IDC. Furthermore, due to a distortion on the leg sensors, there is $70N$ force reading difference between the leg forces. The mismatch between the leg sensors causes the system to be more susceptible to roll oscillations along with bounces and contact losses on the left leg during landing.
\section{Conclusion}
In this work, we analyzed and experimentally compared two different whole-body control formulations for humanoid robots: inverse dynamics whole-body control and passivity-based whole-body control. These two controllers separate from each other as ID-WBC is formulated in task acceleration space, whereas the PB-WBC controller is in task force space. We divide the comparison conclusion into three subtopics: tuning, formulation, and impact robustness.
\subsection{Ease of Tuning}
As the ID-WBC is formulated in task acceleration space, the disturbances and control actions are scaled by the inertia mapping of the system, resulting in different behaviors in different tasks depending on the inertia values. For this controller, the same amount of disturbance results in different magnitudes of errors inversely proportional to the magnitude of inertia. For example, joint friction results in a higher steady-state error in ankle orientation control than the CoM position control, as the former has much smaller task inertia. The same effect is also valid for PD control gain actions in case of disturbance rejection. As the PD control is applied in the acceleration space and the acceleration commands are scaled by the inertia, low-inertia task controls must inherit much higher control gains than the high-inertia task controls to overcome the same amount of disturbance. The P control gain for the foot position control goes as high as $4\times 10^3$, whereas the CoM height control task required only $150$. The same gains go as high as $21 \times 10^3$ for foot orientation control, resulting in high variation of parameters for different tasks. The PB-WBC, on the other hand, formulates the errors and control actions in the same space, resulting in a more uniform behavior among different tasks independent of their inertia. Even though it is possible to obtain similar performance from these two controllers, which is shown analytically and experimentally, PB-WBC appears to be more intuitive and easier to tune. The ID-WBC requires knowledge of the inertia and scales much differently between different tasks and amounts of disturbances. Finally, as the PB-WBC is formulated in task force space, it is a more natural implementation for interaction tasks with certain forces or impedance.
\subsection{Formulation}
The ID-WBC has a straightforward and modular formulation. The task formulations are collected in a QP in a modular way without any Jacobian invertibility condition, as it does not require transformation from the task to joint space. Consequently, ID-WBC works with any set of tasks independently, whether they are under-, well-, or over-determined. The PB-WBC, on the other hand, appears to be more involved in formulating and implementing. It requires mapping from the desired task space states back to joint space, which is easy to handle for under- and well-defined systems but not straightforward for over-determined systems.
\subsection{Robustness Against Landing and Impacts}
It is shown that there exists a set of control gains such that both control systems behave similarly. In the case of jumping, both systems managed to jump similar heights with small differences that mostly originate from differences in the closed-loop dynamics and the overall tuning, as jumping requires more than 40 parameters per whole-body controller to be tuned. On the other hand, in the case of jumping, where a drastic impact takes place on the feet, PB-WBC required additional damping terms to compensate for the violations in the rigid contact assumption that is mostly disturbed due to the leg non-uniformities described earlier. Even though both control systems suffer from leg force non-uniformities, we observe that the PB-WBC suffers comparably more. The reason for that might be the disturbance scaling appearing in the IDC described in \eqref{idc-actual}. As the CoM height, torso, and leg control tasks are high-inertia tasks and the disturbance is scaled by the inverse inertia, the non-uniformity disturbance appears comparably less in IDC. The supplemental video shows that both control systems struggle to stabilize the left leg during landing. We predict that, in case of no discrepancy in the legs, no additional damping would be needed for the PB-WBC, too, as it visually appears to land properly.
\bibliographystyle{IEEEtran}
\bibliography{references}
\end{document}